# On Discarding, Caching, and Recalling Samples in Active Learning

**Ashish Kapoor and Eric Horvitz**
Microsoft Research
1 Microsoft Way
Redmond, WA 98052

## Abstract

We address challenges of active learning under scarce informational resources in non-stationary environments. In real-world settings, data labeled and integrated into a predictive model may become invalid over time. However, the data can become informative again with switches in context and such changes may indicate unmodeled cyclic or other temporal dynamics. We explore principles for discarding, caching, and recalling labeled data points in active learning based on computations of value of information. We review key concepts and study the value of the methods via investigations of predictive performance and costs of acquiring data for simulated and real-world data sets.

## 1 INTRODUCTION

We have been pursuing the challenge of developing systems with the ability to adapt continuously to their environment over the course of their lifetimes. Such lifelong learning systems typically must allocate resources to the collection and labeling of data. Acquiring labels about states of interest that are unavailable in the wild can be costly but important for building predictive models. We explore the use of an active-learning methodology that balances the value and costs of collecting and integrating data. Most work to date on active learning assumes a *pool-based* setting where the set of labeled and unlabeled data are provided and the algorithm selects cases from the pool to query. *Stream-based* learning resonates more deeply with the goals of autonomous lifelong learning. In stream-based settings, the learner sees a series of unlabeled data points and continues to make decisions about whether to query for missing labels. In distinction to pool-based scenarios, learners in stream-based settings may not have complete information about the underlying data distribution. Further, in many dynamic environments, data observed in the past can become outdated due to unmodeled changes in the world. Given incomplete information and dynamic environments, it may be important for a learning system to employ explicit machinery for reviewing and identifying the validity of labeled data points previously encountered and to remove such cases from consideration. We also wish to give a system the ability to reconsider data that was previously removed from consideration.

We shall focus on the example of a system that learns, via the monitoring of computer activity and context, to infer the cost of interrupting a computer user about incoming messages. The system, named BusyBody (Horvitz et al. 2004), was developed originally without active learning machinery. In its initial fielding, the prototype learned probabilistic graphical models from a library of cases collected during a training phase. During training, users of the system are probed at random times, via a pop-up dialog and audio herald, where they are asked to estimate the cost associated with an interruption. The system accrues a growing case library for learning models to predict the cost of interruption by associating the labels with a large vector of evidence about current and recent computer activity, and such contextual cues as time of day and day of week, calendar information, acoustical information, and wireless signals. In earlier work (Kapoor and Horvitz 2007), we extended BusyBody by introducing active learning that guides experience sampling according to a decision-theoretic policy. The methods centered on decisions about probing for labels based on current activity and context. In this paper, we present methods that broaden the active learning methodology to include decisions about discarding and recalling data. The extensions can endow a system with the ability to adapt to unmodeled dynamics that might be encountered in a domain over time. Unexplained non-stationarity is founded on evidential incompleteness in the sensing and learning apparatus. Important distinctions that might help a system to model the dynamics are exogenous and unavailable to the learner. As examples, a learning system that does not have access to variables for recognizing the day of week or recurrent appointments might find itself faced with mystifying non-stationarity. For example, a lesioned Busybody that has lost the ability to sense such distinctions might find models perform poorly in some settings for unknown reasons. Moving beyond such salient examples, subtle incompleteness in modeling capabilities might often lead to significant performance losses in attempts to learn within dynamic contexts.

We shall present the use of the *value of forgetting* and the *value of recalling* in active learning as a means for increasing the robustness of learning and reasoning in stream-centric settings. We demonstrate how we can enhance stream-based active learning with a method that continues to make decisions about removing data points from the active set, caching removed cases for potential future use, and reconsidering the cached cases for reuse. The approach centers on estimating the expected value of including data in the case library using value-of-information (VOI) computations. The method provides stream-based learners with the ability to use their current knowledge to make decisions about whether (1) to pay the price of probing for the label of newly observed data, (2) to "forget" and cache the data that had been probed earlier and (3) to "remember" older cached cases in learning revised predictive models.

We build on related work on decision-theoretic active learning that had been employed solely for pool-based situations (Kapoor et al. 2007). In the prior research, the costs of labeling cases and the cost of misclassification (potentially asymmetric) are considered in parallel. The VOI associated with labeling a previously unlabeled instance is estimated for all unlabeled cases and the cases with the highest VOI are queried for labels. Here, we address the challenge of stream-based active learning in a dynamic environment by computing the *expected value of probing* (VOP) when encountered with a new unlabeled instance. Intuitively, VOP is the expected utility of acquiring a label for a new instance from a human or oracle. We also consider methods for identifying when previously queried cases conflict with the current situation. We explore the benefits of enabling the learner to forget and cache previously labeled cases via the computation of the *expected value of forgetting* (VOF) instances. Specifically, we compute the VOF of labeled cases by computing the expected reduction in the cost of misclassification with forgetting the cases in the current context. Similarly, we consider the possibility that cases previously cached could be useful in the current context by computing the *expected value of recall* (VOR). VOR of cached cases is defined as the expected reduction in the cost of misclassification given reconsideration of the cases. In general, computing VOP, VOF, and VOR is difficult. We derive an efficient algorithm to estimate these quantities. The algorithm exploits the structure of Expectation Propagation (EP) (Minka 2001) for classification with Gaussian Processes. We demonstrate how these quantities can be estimated efficiently via the EP procedure.

The rest of the paper is organized as follows: We first review representative related research in active learning. Then, we present a lifelong learning system that harnesses principles of active learning using VOP to seek labels for new cases, VOF to decide whether labeled cases should be removed and cached, and VOR to determine the cached cases that should be returned to the active set. We show how the structure of Gaussian Process classification can be used to derive efficient methods to estimate VOP, VOF, and VOR. After laying out core concepts, we demonstrate the effectiveness of the methods as an extension to the baseline BusyBody system.

## 2　RELATED RESEARCH

Much of the work on active learning to date has focused on pool-based methods. Various heuristics have been employed as criteria for active learning. Within the Gaussian Process framework, the expected informativeness of an unlabeled data point (Mackay 1992, Lawrence et al. 2002) has been popular. For Support Vector Machines (SVMs), Tong and Koller 2000 explored the criterion of minimizing the version space to select the unlabeled points to query. Roy and McCallum 2001 showed how expected reduction in misclassification can be used for active learning with the naive Bayes classifier. Other pool-based active learning methods have been based on combining active learning and semi-supervised learning (McCallum and Nigam 1998, Muslea et al. 2002, Zhu et al. 2003).

Stream-based active learning has been less explored than active learning in pool-based scenarios. Many of the existing approaches can be viewed as adaptations of pool-based strategies to stream-based scenarios. As an example, one approach to stream-based active learning relies on the selection of unlabeled points for which the existing classification is most uncertain (Lewis and Gale 1994). In another approach, researchers have adapted methods that consider the disagreement of a committee of classifiers to the stream-based scenario (Freund et al. 1997). Similarly, stream-based active learning for linear classifiers has been proposed (Cesa-Bianchi et al. 2003, Dasgupta et al. 2005).

All of these methods address the challenge of adding points to the active set; none of the methods tackles the issue of eliminating outdated data that may become irrelevant or erroneous and recalling older data that might become relevant. Further, none of these methods are targeted at optimizing the criterion upon which the system is ultimately evaluated.

There has also been related work in *online* supervised classification. Recent work has explored the removal of training points under budget constraints (Crammer et al. 2003, Dekel et al. 2006).

## 3 THE ACTIVE LEARNING FRAMEWORK

In earlier work on decision-theoretic active learning, we computed the value of probing for labels of cases (Kapoor et al. 2007, Kapoor and Horvitz 2007). We now provide details about broadening the use of value of information to forgetting and recalling cases. We employ a decision-theoretic strategy to derive an active learning criterion that captures the costs of operating a lifelong learner in an environment over time where the policies are aimed at long-term optimization over the expected lifetime of the system.

Table 1 presents the high-level procedure for executing the proposed methodology for active learning. The approach has three major components. On encountering an unlabeled data point from the stream, we perform a *seek cycle*, a *cache cycle*, and a *recall cycle*. In the seek cycle, VOP computations determine if the learner should pay the price of probing for the label of an observed data point. The cache cycle computes VOF for all of the cases being considered in the model (the set $L$ of *active* labeled cases), and determines if points should be discarded from the active set and cached for the later use. Additionally, the algorithm maintains a buffer $B$ of a finite length $S_{buff}$, holding the most recently encountered cases. The buffer of cases is used to capture non-stationarity. Note that the buffer $B$ is different from the set of active labeled points $L$ that is used to train the classification model. The buffer $B$ provides a means for estimating the current underlying distribution of the observations used in the algorithm to compute the VOP, VOF, and VOR within the current context. Further, the VOP computation also considers an *optimization horizon*, $k_{horiz}$ that defines the duration of system usage considered in the long-term lifelong learning optimization (Kapoor and Horvitz 2007). $k_{horiz}$ refers to the number of unobserved points that will be seen by the system and determines the tradeoff between the acute cost of a probe and the long-term benefits associated with the expected improvements of the system's performance associated with refining the model using the additional labels. We describe the VOP, VOF, and VOR components in greater detail following a discussion of several assumptions we wish to make explicit.

First, for tractability, we perform myopic computations, where we determine VOP, VOF and VOR in

Table 1: Decision-theoretic active learning

---

Input Data Stream: $\mathcal{X}_T = \{\mathbf{x}_1, ..\mathbf{x}_{t-1}, \mathbf{x}_t, \mathbf{x}_{t+1}, .., \mathbf{x}_T\}$
Initial Classifier: $\mathbf{w_0}$
Maximum size of the buffer: $S_{buff}$
Size of Horizon: $k_{horiz}$

---

set of active points $L = \{\}$, cache $C = \{\}$
　　and buffer $B = \{\}$

for $t = 1, .., T$
　　Observe the data $\mathbf{x}_t$
　　$B = B \cup \mathbf{x}_t$
　　if size$(B) > S_{buff}$
　　　　discard the oldest point
　　end

　　%**Seek cycle**: *pursuing new labels*
　　If VOP$(\mathbf{x}_t, \mathbf{w}_{t-1}, k_{horiz}) > 0$
　　　　add to active set $L = L \cup \mathbf{x}_t$
　　end
　　update the classifier: $\mathbf{w}_t$

　　%**Cache cycle**: *forgetting & storing labeled cases*
　　for all the labeled points $\mathbf{x}_l$ in $L$
　　　　if VOF$(\mathbf{x}_l, \mathbf{w}_t) > 0$
　　　　　　remove from active set $L = L - \mathbf{x}_l$
　　　　　　add to cache $C = C \cup \mathbf{x}_l$
　　　　end
　　end
　　update the classifier: $\mathbf{w}_t$

　　%**Recall cycle**: *remembering discarded cases*
　　for all the cached labeled points $\mathbf{x}_l$ in $C$
　　　　if VOR$(\mathbf{x}_l, \mathbf{w}_t) > 0$
　　　　　　add to active set $L = L \cup \mathbf{x}_l$
　　　　　　remove from cache $C = C - \mathbf{x}_l$
　　　　end
　　end
　　update the classifier: $\mathbf{w}_t$
end

---

separate phases of analysis that consider points one at a time. Further, for simplicity we present the work with examples involving binary (two class) discrimination problems.

For discussion purposes, we focus on *linear classifiers*, parameterized as $\mathbf{w}$ and classify test points $\mathbf{x}$ according to: sign$(f(\mathbf{x}))$, where $f(\mathbf{x}) = \mathbf{w}^T\mathbf{x}$. Let the set of labeled data points be denoted by $\mathcal{X}_L = \{\mathbf{x}_1, .., \mathbf{x}_L\}$, with class labels $\mathcal{T}_L = \{t_1, .., t_L\}$, where $t_i \in \{1, -1\}$, then the lifelong learner learns the parameters $\mathbf{w}$. Further, we define the risk matrix $\mathbf{R} = [R_{ij}] \in I\!\!R^{2\times 2}$, where $R_{ij}$ denotes the cost or *risk* associated with classifying a data point belonging to class $i$ as $j$. We use the index 2 to denote the class -1. We assume that the diagonal elements of $\mathbf{R}$ are zero, asserting that the correct classification incurs no cost. Further, let $C_i$ denote the cost of querying to obtain the class label of $\mathbf{x}_i$. We assume that the costs of querying $C_i$ and the risks $R_{12}$ and $R_{21}$ are measured with the same currency, *e.g.*, dollars. Further, given a classifier $\mathbf{w}$, the total cost of misclassification or *risk* $J$ on all of the data points in the buffer $B$ can be written as:

$$J = \sum_{i \in B} R_{12} \cdot \mathbf{1}_{[\mathbf{w}^T\mathbf{x}_i<0]} p_i^* + R_{21} \cdot \mathbf{1}_{[\mathbf{w}^T\mathbf{x}_i>0]} (1-p_i^*)$$

Here, $\mathbf{1}_{[\cdot]}$ is the indicator function and $p_i^* = p(t_i = 1|\mathbf{x}_i)$ is the *true* conditional density of the class label given the data point. As we do not have the true conditional distribution, we make approximations. Note, that the current state of the classifier captures the learner's belief about the distribution of the label given the data point. Thus, if available, we can use this predictive distribution to approximate $p_i^*$. Specifically, let $p_i$ denote the probability that the point $\mathbf{x}_i$ belongs to class +1 according to the classification model, i.e., $p_i = p(\text{sign}(f(\mathbf{x}_i)) = 1|\mathbf{x}_i)$. Then, we can use $p_i^* \approx p_i$. Note, that $p_i$ is the predictive distribution and its availability will depend on the base-level classification technique being used. Predictive distributions are available for Gaussian Process classification and other probabilistic classifiers, including probabilistic mappings of outputs of SVMs (Platt 2000).

### 3.1 SEEK CYCLE: PURSUING NEW LABELS

The *seek cycle* is based on the methods developed in prior research on pool-based active learning (Kapoor et al. 2007). We propose modifications to the earlier approach to adapt it to the stream-based setting. Consider the prospect that knowing the label of the currently observed unlabeled point can reduce the total misclassification risk for actions taken in the future. However, labels are acquired at a price. The difference in the reduction in the risk and the cost of acquiring a new label is the expected value of information for learning that label.

Formally, given the classifier $\mathbf{w}_{t-1}$ available at time step $t$ and the buffer $B$, we compute the VOP of the unobserved label, corresponding to the current data $\mathbf{x}_t$, as the difference in the reduction in misclassification risk over the optimization horizon $k_{horiz}$ and the cost of obtaining the label:

$$\text{VOP}(\mathbf{x}_t, \mathbf{w}_{t-1}, k_{horiz}) = k_{horiz} \cdot \Delta - C_t \quad (1)$$

Here, $\Delta$ is the average reduction in the misclassification risk and we estimate it with the empirical mean as: $\Delta = \frac{J - J^t}{|B|}$, where $J^t$ denotes the expected risk on the points in the buffer $B$ if the current data $\mathbf{x}_t$ was labeled. We point out that the computation of VOP assumes *local* stationarity within the current context which is modeled by the buffer $B$.

Note that we do not know the label of $\mathbf{x}_t$ when performing the VOP computation. Thus, $J^t$ is computed by taking expectations over the label:

$$J^t = J^{t,+} p_t + J^{t,-}(1-p_t)$$

$J^{t,+}$ ($J^{t,-}$) correspond to the misclassification risk when $\mathbf{x}_t$ is labeled as +1 (-1). The VOP quantifies the gain in utility that can be obtained by obtaining a new label. Thus, our strategy is to seek a label for an unlabeled case whenever VOP $\geq 0$. Note, that a large $k_{horiz}$ will typically weight the system toward seeking labels for cases early on, while decreasing $k$ increases the system's reluctance to ask for supervision.

### 3.2 CACHE CYCLE: FORGETTING AND STORING LABELED CASES

The intuition behind the cache cycle is that removing points from the case library can lead to a significant change in the decision surface. Further, in dynamic scenarios, such a change in the decision surface can significantly reduce the misclassification risk in the current context. The current context is represented by data in the buffer $B$ containing the most recent points. The reduction in the misclassification risk on $B$ with the discarding of active points is the expected value of forgetting.

Specifically, given the current classifier $\mathbf{w}_t$ at the time step $t$ and the buffer $B$, we can compute the VOF of an active data point $\mathbf{x}_i$ as the reduction in misclassification risk due to the removal:

$$\text{VOF}(\mathbf{x}_i, \mathbf{w}_t) = J - J^{/i} \quad (2)$$

Here, $J^{/i}$ corresponds to the misclassification risk on $B$ when $\mathbf{x}_i$ is discarded. VOF quantifies the utility gain that can be obtained by discarding a point from the active set. We remove all of the cases from the active set $L$ with VOF $\geq 0$. These points are cached for reconsideration and potential reuse in the future.

Note that the naive computation of VOF requires the training of classifiers for all possible leave-one-out instantiations of the active set. Such computations can be expensive in the general case. We employ Gaussian Process classification and exploit the structure of the problem to gain significant computational advantages.

### 3.3 RECALL CYCLE: REMEMBERING DISCARDED CASES

Caching the discarded cases allows the system to reuse them when appropriate. In dynamic environments, these cached data points can again become relevant based on aspects of context exogenous to the distinctions represented explicitly in the modeling machinery. As we already paid the price to acquire labels for these points, it is less expensive to re-incorporate them than to probe for an unlabeled instance. Given the context represented by data in the buffer $B$, we compute the reduction in the misclassification risk on reusing a cached case. This reduction is the expected value of recalling.

Formally, given the current classifier $\mathbf{w}_t$ at the time step $t$ and the buffer $B$, we compute the VOR of a cached case $\mathbf{x}_i$ as the reduction in misclassification risk with the inclusion of the case:

$$\text{VOR}(\mathbf{x}_i, \mathbf{w}_t) = J - J^{\cup i} \quad (3)$$

Here, $J^{\cup i}$ corresponds to the misclassification risk on $B$ when $\mathbf{x}_i$ is reintroduced in L. The VOR quantifies the gain in utility that can be obtained by reincorporating a cached point into the active set. We reintroduce all of the cases from the cache with VOR $\geq 0$.

## 4 COMPUTATIONAL ASPECTS

We have employed Gaussian Process (GP) classifiers within the proposed active-learning methodology. There are two major advantages with this choice: first, we directly model the predictive conditional distribution $p(t|\mathbf{x})$, making it easy to compute the actual conditional probabilities without any calibrations or post-processing. Second, Expectation Propagation (EP) (Minka 2001), an approximate inference algorithm we use for GP classification, enables the efficient computations of VOP, VOF, and VOR.

As a brief overview, the GP methodology provides a Bayesian interpretation of classification. With the approach, the goal is to infer the posterior distribution over the set of possible classifiers given a training set:

$$p(\mathbf{w}|\mathcal{X}_L, \mathcal{T}_L) = p(\mathbf{w}) \prod_{i \in L} p(t_i|\mathbf{w}, \mathbf{x}_i) \quad (4)$$

Here, $p(\mathbf{w})$ corresponds to the prior distribution over the classifiers and is selected typically so as to prefer parameters $\mathbf{w}$ that have a small norm. Specifically, we assume a spherical Gaussian prior on the weights: $\mathbf{w} \sim \mathcal{N}(0, \mathbf{I})$. The prior imposes a smoothness constraint and acts as a regularizer such that it gives higher probability to the labelings that respect the similarity between the data points. The likelihood terms $p(t_i|\mathbf{w}, \mathbf{x}_i)$ incorporate the information from the labeled data and different forms of distributions can be selected. A popular choice is the probit likelihood: $p(t|\mathbf{w}, \mathbf{x}) = \Psi(t \cdot \mathbf{w}^T \mathbf{x})$. Here, $\Psi(\cdot)$ denotes the cumulative density function of the standard normal distribution. The posterior prefers those parameters that have small norm and that are consistent with the data.

Computing the posterior, $p(\mathbf{w}|\mathcal{X}, \mathcal{T})$, is non-trivial and approximate inference techniques such as Assumed Density Filtering (ADF) or Expectation Propagation (EP) are required. The idea behind ADF is to approximate the posterior $p(\mathbf{w}|\mathcal{X}_L, \mathcal{T}_L)$ as a Gaussian distribution, i.e. $p(\mathbf{w}|\mathcal{X}_L, \mathcal{T}_L) \approx \mathcal{N}(\bar{\mathbf{w}}, \Sigma_\mathbf{w})$. Similarly, EP is another approximate inference technique and a generalization of ADF, where the approximation obtained from ADF is refined using an iterative message passing scheme. We refer readers to Minka 2001 for details.

Given the approximate posterior $p(\mathbf{w}|\mathcal{X}, \mathcal{T}) \sim \mathcal{N}(\bar{\mathbf{w}}, \Sigma_\mathbf{w})$, a frequent practice is to choose the mean $\bar{\mathbf{w}}$ of the distribution as the point classifier. The mean, which is also called the *Bayes point*, classifies a test point according to: $\text{sign}(\bar{\mathbf{w}}^T \mathbf{x})$. The non-linear case can be generalized by using the kernel trick, in which the data is projected into a higher dimensional space to make it separable. Note, that the predictive distribution $p(\text{sign}(f(\mathbf{x}))|\mathbf{x})$ is given by:

$$p(\text{sign}(f(\mathbf{x})) = \mathbf{1}|\mathbf{x}) = \Psi\left(\frac{\bar{\mathbf{w}}^T \mathbf{x}}{\sqrt{\mathbf{x}^T \Sigma_\mathbf{w} \mathbf{x} + 1}}\right) \quad (5)$$

Unlike other classifiers, the GP classification models the predictive conditional distribution $p(t|\mathbf{x})$, making it easy to compute the actual conditional probabilities without any calibrations or post-processing. This predictive distribution is used to compute misclassification risks in the active learning framework in computing VOP, VOF, and VOR.

Important byproducts of using EP or ADF are the Gaussian approximations of the likelihood terms. Specifically, we have:

$$p(t_i|\mathbf{w}, \mathbf{x}_i) \approx q_i = s_i \exp\left[-\frac{1}{2v_i}(\mathbf{w}^T \mathbf{x}_i \cdot t_i - m_i)^2\right]$$

Here $s_i, m_i$ and $v_i$ are terms computed by EP for approximate inference and together they satisfy:

$$p(\mathbf{w}|\mathcal{X}_L, \mathcal{T}_L) \approx \mathcal{N}(\bar{\mathbf{w}}, \Sigma_\mathbf{w}) = p(\mathbf{w}) \prod_{i \in L} q_i \quad (6)$$

We note that computing the leave-one-out posterior is just the removal of one Gaussian terms corresponding to the likelihoods from the final posterior $p(\mathbf{w}|\mathcal{X}_L, \mathcal{T}_L)$, which is a Gaussian distribution in itself. Formally, we need to remove the Gaussian $q_j$ from the original posterior to compute $p^{/j}(\mathbf{w}|\mathcal{X}_L, \mathcal{T}_L)$, the posterior distribution obtained by leaving out the $j^{th}$ data point from $L$. Specifically, the leave-one-out mean $\bar{\mathbf{w}}^{/j}$ and the variance $\Sigma_\mathbf{w}^{/j}$ can be written as:

$$\Sigma_\mathbf{w}^{/j} = \Sigma_\mathbf{w} + (\Sigma_\mathbf{w} \mathbf{x}_j)(v_j - \mathbf{x}_j^T \Sigma_\mathbf{w} \mathbf{x}_j)^{-1}(\Sigma_\mathbf{w} \mathbf{x}_j)^T$$
$$\bar{\mathbf{w}}^{/j} = \bar{\mathbf{w}} + (\Sigma_\mathbf{w}^{/j} \mathbf{x}_j) v_j^{-1} (\bar{\mathbf{w}}^T \mathbf{x}_j - m_j)$$

Thus, we have $p^{/j}(\mathbf{w}|\mathcal{X}_L, \mathcal{T}_L) = \mathcal{N}(\bar{\mathbf{w}}^{/j}, \Sigma_\mathbf{w}^{/j})$. We note that these computations are performed in the course of Expectation Propagation (see Minka 2001). That is, when using Expectation Propagation, the leave-one-out posteriors needed for the VOF analysis are provided for free from EP. We can compute VOF without retraining the classification system, resulting in a significant computational savings.

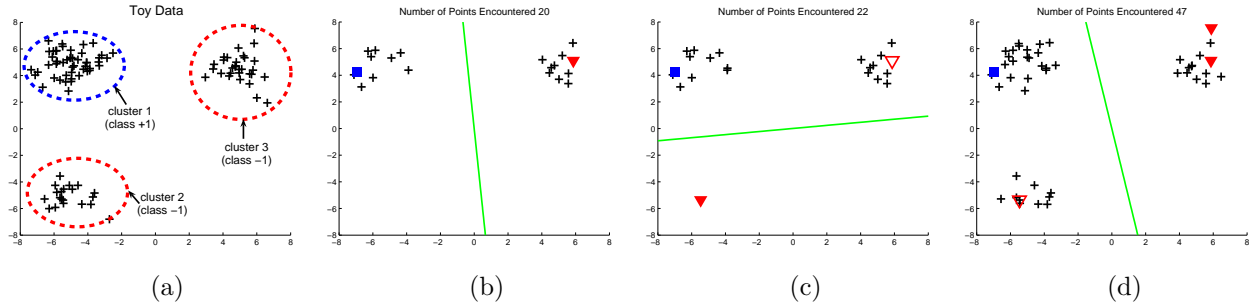

Figure 1: (a) The cluster data set where the points from cluster 1 belongs to class +1 and the rest to class -1 (see text). State of the active learning system after encountering (b) 20 points, (c) 22 points and (d) 47 points. The '+' symbol, square and triangle denote an unlabeled data, data labeled as +1 and data labeled as -1 respectively. The filled (unfilled) squares and triangles correspond to the labeled data points in the active set (cache).

Computing VOP and VOR can be expensive as the computational complexity for EP is $O(n^3)$, where $n$ is the size of labeled training set. In the proposed method, we must compute VOP for every new unlabeled case, requiring us to perform EP twice for every new point under consideration. Similarly, we have to retrain the classifier for every data point in the cache to compute VOR. A faster alternative is to use ADF for approximating the new posterior over the classifier rather than doing full EP. Specifically, to compute the new posterior $p^{j,+}(\mathbf{w}|\mathcal{X}_{L\cup j}, \{\mathcal{T}_L \cup +1\})$ we can compute the Gaussian projection of the old posterior multiplied by the likelihood term for the $j^{th}$ data point. That is: $p^{j,+}(\mathbf{w}|\mathcal{X}_{L\cup j}, \{\mathcal{T}_L \cup +1\}) \approx \mathcal{N}(\bar{\mathbf{w}}^{j,+}, \Sigma_{\mathbf{w}}^{j,+})$, where $\bar{\mathbf{w}}^{j,+}$ and $\Sigma_{\mathbf{w}}^{j,+}$ are respectively the mean and the covariance of $p(\mathbf{w}|\mathcal{X}_L, \mathcal{T}_L) \cdot \Psi(1 \cdot \mathbf{w}^T \mathbf{x}_j)$. This is equivalent to performing ADF starting with the old posterior $p(\mathbf{w}|\mathcal{X}_L, \mathcal{T}_L)$ and incorporating the likelihood term $\Psi(1 \cdot \mathbf{w}^T \mathbf{x}_j)$ and does not require $O(n^3)$ operations to compute VOI for every unlabeled data point. We can use similar computations to approximate $p^{j,-}(\mathbf{w}|\mathcal{X}_{L\cup j}, \{\mathcal{T}_L \cup -1\})$.

As mentioned earlier, we perform myopic computations of VOP, VOF, and VOR. However, rather than removing and recalling one point at a time, we choose the policy of forgetting and recalling all the points from $L$ that have VOF > 0 and VOR > 0 respectively with a goal of alleviating any ordering effects. We expect that relaxing the myopia will improve the performance. We are pursuing such generalizations of the methodology.

## 5 EXPERIMENTS

We tested the effectiveness of the methods on a synthetic data set and on the real-world task of classifying the cost of interrupting a user. All of the experiments described were carried out with a linear GP classifier as the underlying predictive model. Further, we set $S_{buff} = 5$ and $k_{horiz}$ to be equal to the length of the sequence. We compare the proposed learning scheme with a method that only probes but cannot discard, cache, and recall cases. We also compare the method with two heuristic policies. First, we consider the policy of randomly selecting cases to query the user. The other scheme selects cases on which the classification model is most uncertain. Specifically, with this policy, the system probes for labels if $0.3 \leq p(t|\mathbf{x}) \leq 0.7$.

### 5.1 DATA SETS

We performed experiments with the following data sets:

**Cluster Data:** This is a synthetic data set as shown in Figure 1(a). Here the data are generated sequentially forming 3 clusters, where the points coming from cluster 1 belong to class +1 and the rest to −1. Further, the points are generated in a sequence such that the data generation process switches between clusters 2 and 3 after every 20 points. Thus, there is a temporal characteristic to this data.

**BusyBody Data:** This is the real-world data collected by from Busybody users. BusyBody logs desktop events including keyboard and mouse activity, windows in focus, recent sequences of applications and window titles, and high-level statistics about the rates of switching among applications and windows. The system also considers several classes of contextual variables, including the time of day and day of week, the name of the computer being used, the presence and properties of meetings drawn from an electronic calendar, and wireless signals. The system also employs a conversation-detection system, using a module that detects signals in the human-voice range of the audio spectrum. Our aim is to learn a mapping from the logged data to the cost of interrupting the user. In the tests, we performed simulations on temporal data previously collected by the BusyBody system for two subjects. The first user is a program manager and the other is a software developer. The data for each subject contains two weeks of desktop activity as well as the busy/not-busy tags that had been collected by the original BusyBody system, using its legacy random probe policy. We only consider data in the sequence that were labeled by such assessments.

Table 2: Comparison: Cluster Data

| METHOD | PROBES | COST | ACCURACY |
|---|---|---|---|
| VOP+VOF+VOR | **5** | **8** | **96.84%** |
| VOP | 9 | 14 | 94.51% |
| Random (p=0.05) | 8 | 27 | 79.35% |
| Most Uncertain | 10 | 23 | 85.56% |

Table 3: Comparison: Program Manager Data

| METHOD | PROBES | COST | ACCURACY |
|---|---|---|---|
| VOP+VOF+VOR | **35** | **172** | **68.29%** |
| VOP | 109 | 316 | 66.19% |
| Random (p=0.1) | 50 | 354 | 57.75% |
| Most Uncertain | 258 | 531 | 63.26% |

Table 4: Comparison: Developer Data

| METHOD | PROBES | COST | ACCURACY |
|---|---|---|---|
| VOP+VOF+VOR | 12 | **34** | 85.59% |
| VOP | 39 | 77 | **89.01%** |
| Random (p=0.05) | **8** | 48 | 78.69% |
| Most Uncertain | 55 | 127 | 78.67% |

Note that we evaluate the proposed system on all of the points where the learner chose *not* to probe. Further, in the evaluation, the system employs the classification model trained using the data seen *up to* the point being tested. Thus, we can observe and characterize the performance of the system as it is evolving. We assume an asymmetric loss where the cost of misclassifying the *busy* state as *not busy* is twice as expensive as misclassifying *not busy* as *busy*. Specifically we have $r_{12} = 2$ USD and $r_{21} = 1$ USD. Similarly we assume that 2 USD is the cost of probing the user for a label when the user is busy and 1 USD when the user is not busy. For the cluster data, we assume a symmetric loss and a cost of probing equal to 1 USD.

## 5.2 RESULTS AND DISCUSSION

Figure 1(b), (c), and (d) demonstrate graphically the various phases in active learning for the cluster data. The black '+' symbols denote the unlabeled data and the square and the triangles correspond to the data that was probed. Further, the filled triangles and squares correspond to the data points in the active set, whereas the unfilled shapes denote that the data points are in the cache. Figure 1(b) shows the state of the system after encountering 20 data points. The state significantly changes after encountering 22 data points (Figure 1(c)) where the system correctly decides to put one of the active points in the cache; thus, correctly classifying the other unlabeled points belonging to the cluster 2. Similarly, when the system starts encountering more points from cluster 3 (Figure 1(d)), it decides to discard the labeled point from the cluster 2 and retrieve the earlier cached point. This toy example demonstrates the effectiveness of the framework. Table 2 shows the cost incurred and the accuracy achieved by the various methods. The proposed method (VOP+VOF+VOR) achieves the highest accuracy with the lowest number of probes and the cost. Note that the total cost includes the loss associated with misclassification and the cost of interruptions from the probes themselves.

Tables 3 and 4 show the recognition accuracy on the unlabeled points and the total cost incurred for the BusyBody data sets. The lifelong learning method that has the capability to forget, cache, and recall (VOP+VOF+VOR) cases significantly outperformed the systems that only used VOP and the heuristic policies in terms of the total cost incurred. Further the VOP+VOF+VOR scheme beats the other approaches in terms of performance accuracy, except for the case of the developer data where the scheme that just employs VOP performs slightly better albeit, with a high cost based in the large number of probes. We found that, despite the simplicity of the linear classifier, the ability to discard and recall cases is associated with a significant improvement in the performance of the system. We believe that the overall boosts in performance is based in the value of the proposed machinery to enable the learning system to continue to adapt in environments in the common situation where modeling distinctions and machinery are blind to important distinctions that remain unmodeled. Executing the lifelong learning scheme resulted in overall accuracies of 68.29% for the program manager and 85.59% for the developer. The program manager was probed 35 times and the developer was probed 12 times. The VOP+VOF+VOR scheme provides significant gains in recognition accuracy at very little additional computation costs and with fewer probes (and thus fewer interruptions) to the user. Figure 2 plots the cost incurred as the system sees progressively more labels. The graph highlights the benefits of employing machinery for forgetting and caching irrelevant data points and recalling older cached points in a principled manner.

## 6 CONCLUSION

We presented a methodology for labeling, removing, and reincorporating cases for learning in dynamic environments. We showed how the costs of misclassification and of obtaining labels can be used to quantify the value of probing for labels, and the value of forgetting and recalling cases in stream-centric learning. We described a tractable approach to such decision making that exploits the structure of Expectation Propagation. We tested key concepts via a set of experiments. We found that the methods were advanta-geous for the test domain, on learning and continuing to update models that predict the cost of interrupting computer users. The studies highlight the promise of executing VOP, VOF, and VOR cycles in lifelong learning scenarios. We are pursuing several challenges and opportunities with learning and forgetting in active learning, including the extension of the analysis to multiple classes, the development of less myopic analyses, optimization of the size of the context buffer, and exploration of automated techniques for identifying and incorporating exogenous variables, such as those that could inform a system about temporal periodicities and subcontexts.

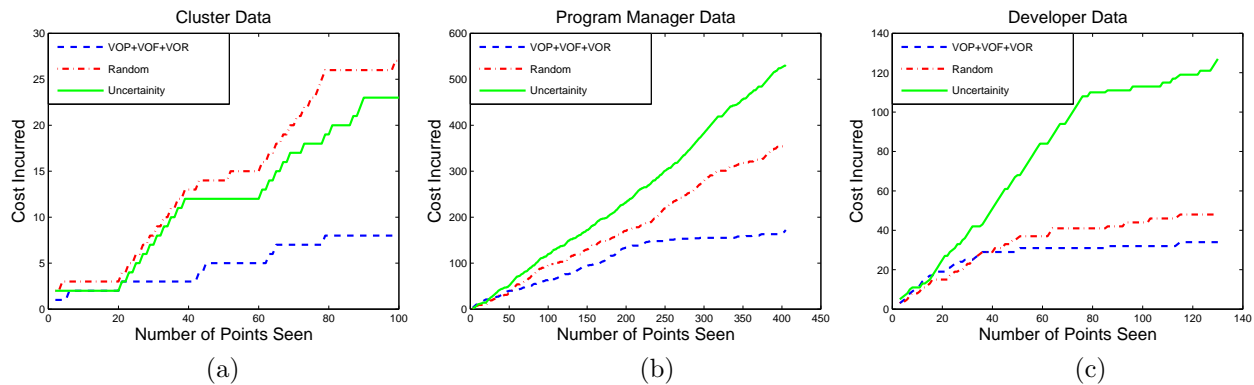

Figure 2: The total cost incurred by the system on the test points as it encounters cases for the (a) cluster data, (b) program manager data and (c) developer data.

## References

N. Cesa-Bianchi, A. Conconi and C. Gentile (2003). Learning probabilistic linear-threshold classifiers via selective sampling. Conference on Learning Theory.

K. Crammer, J. Kandola and Y. Singer (2003). Online classification on a budget. Neural Information Processing Systems.

S. Dasgupta, A. T. Kalai and C. Monteleoni (2005). Analysis of perceptron-based active learning. Conference on Learning Theory.

O. Dekel, S. Shalev-Shwartz and Y. Singer (2006). The Forgetron: A Kernel-Based Perceptron on a Fixed Budget. Neural Information Processing Systems.

Y. Freund, H. S. Seung, E. Shamir and N. Tishby (1997). Selective Sampling Using the Query by Committee Algorithm. Machine Learning Volume 28.

E. Horvitz, J. Apacible and P. Koch (2004). BusyBody: Creating and Fielding Personalized Models of the Cost of Interruption. Conference on Computer Supported Cooperative Work.

A. Kapoor, E. Horvitz and S. Basu (2007). Selective Supervision: Guiding Supervised Learning with Decision-Theoretic Active Learning. International Joint Conference on Artificial Intelligence.

A. Kapoor and E. Horvitz (2007). Principles of Lifelong Learning for Predictive User Modeling. International Conference on User Modeling.

N. Lawrence, M. Seeger and R. Herbrich (2002). Fast Sparse Gaussian Process Method: Informative Vector Machines. Neural Information Processing Systems.

D. D. Lewis and W. A. Gale (1994). A sequential algorithm for training text classifiers. International Conference on Research and Development in Information Retrieval.

D. MacKay (1992). Information-Based Objective Functions for Active Data Selection. Neural Computation Volume 4(4).

A. K. McCallum and K. Nigam (1998). Employing EM in pool-based active learning for text classification. International Conference on Machine Learning.

T. P. Minka (2001). Expectation Propagation for approximate Bayesian inference. Uncertainty in Artificial Intelligence.

I. Muslea, S. Minton and C. A. Knoblock (2002). Active + Semi-supervised Learning = Robust Multi-View Learning. International Conference on Machine Learning.

J. Platt (2000). Probabilities for support vector machines. Advances in Large Margin Classifiers.

N. Roy and A. McCallum (2001). Toward Optimal Active Learning through Sampling Estimation of Error Reduction. International Conference on Machine Learning.

S. Tong and D. Koller (2001). Support Vector Machine Active Learning with Applications to Text Classification. Journal of Machine Learning Research Volume 2.

X. Zhu, J. Lafferty and Z. Ghahramani (2003). Combining Active Learning and Semi-Supervised Learning Using Gaussian Fields and Harmonic Functions. Workshop on The Continuum from Labeled to Unlabeled Data in Machine Learning and Data Mining.